%% file: ID_paper.tex
\documentclass[conference]{IEEEtran}
\usepackage{ifpdf}
\usepackage{pgfplots}
\usepackage{tikz}
\usepackage{subfig}                     % for subfigures
\usepackage{float}
\usepackage{booktabs}                   % nice looking tables (for tables with ONLY horizontal lines)
\usetikzlibrary{calc}

\usepackage[T1]{fontenc}
\usepackage[utf8]{inputenc}
\usepackage[utf8]{inputenc}
\usepackage{pgfplots}
\usepackage{grffile}
\pgfplotsset{compat=newest}
\usetikzlibrary{plotmarks}
\usetikzlibrary{arrows.meta}
\usepgfplotslibrary{patchplots}
\usepackage{graphicx}
\usepackage{adjustbox}
\usepackage{units}

\usepackage{stfloats}
\usepackage{epstopdf}
\usepackage{bm}
\usepackage{amssymb}
\usepackage[abs]{overpic}
\usepgfplotslibrary{external} 
\begin{document}
%	\title{Experimental Study of Body Characteristics Effect on Micro-Doppler-Based Person Identification}
%	\title{A Deep Learning Study of Human Body Characteristics Effect on Micro-Doppler-Based Person Identification}
\title{Person Identification and Body Mass Index:\\ A Deep Learning-Based Study on Micro-Dopplers}
	\author{\IEEEauthorblockN{Sherif Abdulatif\IEEEauthorrefmark{4}\IEEEauthorrefmark{1},
			Fady Aziz\IEEEauthorrefmark{2}\IEEEauthorrefmark{1},
			Karim Armanious\IEEEauthorrefmark{4},
			Bernhard Kleiner\IEEEauthorrefmark{2},
			Bin Yang\IEEEauthorrefmark{4},
			Urs Schneider\IEEEauthorrefmark{2}}
		\IEEEauthorblockA{\IEEEauthorrefmark{4}Institute of Signal Processing and Systems Theory, University of Stuttgart}
		\IEEEauthorblockA{\IEEEauthorrefmark{2}Fraunhofer Institute for Manufacturing Engineering and Automation IPA\\
			Email: sherif.abdulatif@iss.uni-stuttgart.de, fady.aziz@ipa.fraunhofer.de\\
			\IEEEauthorrefmark{1}These authors contributed to this work equally.}}


	\maketitle

	\begin{abstract}
		%\boldmath
		Obtaining a smart surveillance requires a sensing system that can capture accurate and detailed information for the human walking style. The radar micro-Doppler ($\boldsymbol{\mu}$-D) analysis is proved to be a reliable metric for studying human locomotions. Thus, $\boldsymbol{\mu}$-D signatures can be used to identify humans based on their walking styles. Additionally, the signatures contain information about the radar cross section (RCS) of the moving subject. This paper investigates the effect of human body characteristics on human identification based on their $\boldsymbol{\mu}$-D signatures. In our proposed experimental setup, a treadmill is used to collect $\boldsymbol{\mu}$-D signatures of 22 subjects with different genders and body characteristics. Convolutional autoencoders (CAE) are then used to extract the latent space representation from the $\boldsymbol{\mu}$-D signatures. It is then interpreted in two dimensions using t-distributed stochastic neighbor embedding (t-SNE). Our study shows that the body mass index (BMI) has a correlation with the $\boldsymbol{\mu}$-D signature of the walking subject. A 50-layer deep residual network is then trained to identify the walking subject based on the $\boldsymbol{\mu}$-D signature. We achieve an accuracy of 98\% on the test set with high signal-to-noise-ratio (SNR) and 84\% in case of different SNR levels.
	\end{abstract}
	

	% \fi
	\IEEEpeerreviewmaketitle

	% 	\vspace{-0.3cm}
	\section{Introduction}
	\par Studying the human gait is considered a complex process due to the deformability of a moving human body. The human walking gait can be analyzed by measuring the associated temporal and spatial parameters. These spatio-temporal parameters reflect the main characteristics of the human gait motion, e.g. the stride length, the walking velocity and the periods of swing and stance phases. Accordingly, the motion of each body part can be represented by a translational or rotational trajectory. They can then be used to identify the body parts position and orientation over time as proposed in the global human walking model \cite{boulic1990global}. The main parameters used to define the motion trajectories and thus the walking style are the heights of the spine, the thigh and the normalized relative walking velocity.
	
	\par Such models can be used to characterize the motion characteristics of a walking person including the motion behavior of different body parts. However, to obtain an accurate study of the human gait, the body weight is another useful factor. It has been illustrated in \cite{hwang2017effects,blaszczyk2011impact} that the body weight affects the spatio-temporal parameters of the walking gait. Therefore, the walking style can be a useful metric for human identification when accurately evaluated \cite{cutting1977recognizing}. 
	
	\par The human identification and detection play a great importance for many security and medical applications. Different monitoring systems have been designed for gait evaluation depending on the desired application. One commonly used technique is based on evaluating the two main phases of the walking gait cycle, the stance and the swing. The stance phase happens mainly during feet double support on the ground and occupies nearly $60\%$ of the gait cycle. The remaining $40\%$ is occupied by the swing phase which happens when one foot is in total contact with the ground while the other is swinging. To measure the periods of both gait phases, wearable sensors have been used such as the ultrasonic sensors in \cite{qi2016assessment}. Other solutions are based on  motion capture systems to get accurate information of the motion trajectories \cite{abdulatif2017real}. These motion capture systems offer high accuracy in evaluating the motion trajectories of the different body parts by tracking the fixed infrared markers using high resolution cameras. However, the usage of either a capture system or wearable sensors is not practical and not suitable for real-life applications. 
	
	\par Therefore, non-wearable sensors such as radar can be considered as a more efficient alternative system as it can overcome many limitations that other vision-based systems suffer from. For example, radar can be used under poor lighting and harsh environmental conditions as dust and smoke \cite{garcia2014analysis}. Moreover, the micro-Doppler ($\boldsymbol{\mu}$-D) signature presented in \cite{microdopplerBook} has introduced the feasibility of using radar for studying the human gait. Studying the  $\boldsymbol{\mu}$-D signature can also give an evaluation of the micro-motion behavior of different body parts within the gait cycle. Accordingly, numerous applications can be achieved by deploying the $\boldsymbol{\mu}$-D signature analysis. For instance, differentiation between human activities such as walking, running and crawling based on their $\boldsymbol{\mu}$-D signatures \cite{kim2016class,tekeli2016information}. Other studies used the $\boldsymbol{\mu}$-D characteristics to differentiate between humans and other moving targets \cite{van2018micro,abdulatif2018micro}.
	
	\par For realizing an intelligent motion-based surveillance system, a descriptive and detailed locomotion data is required. Due to the sufficient data that can be extracted from the $\boldsymbol{\mu}$-D signatures and the powerful capability of learning approaches, radar-based human identification using deep convolutional neural networks (DCNNs) has been presented in different studies.
	In \cite{Vander2018ID}, authors recorded the $\boldsymbol{\mu}$-D signatures of 5 subjects moving randomly in a room with a \unit[77]{GHz} FMCW radar. Afterwards, they used a 4-layer DCNN followed by a fully connected layer to identify moving subjects with an accuracy of 78.5\% for a delay window of \unit[3]{s}. An accuracy of 100\% for a delay window of \unit[25]{s} is mentioned. A \unit[24]{GHz} CW radar was used in \cite{Cao2018ID} to differentiate between 20 moving subjects based on their corresponding $\boldsymbol{\mu}$-D signatures. An accuracy of 97.1\% on 4 subjects was achieved using AlexNet \cite{krizhevsky2012imagenet}. However, the performance degrades in case of more subjects and reaches 68.9\% for 20 subjects. 
	
	In this paper, a study of the effect of the human body characteristics on their measured $\boldsymbol{\mu}$-D signatures is presented. Moreover, a deeper network with 50 layers is shown to achieve a better subject identification accuracy.
	
	
	%\hfill November 1, 2016
	% 	\vspace{-3mm}
%	\vspace{-2mm}
	\section{Dataset Preparation}
	Our experiment is designed to focus on understanding the effect of human body characteristics on their walking style and the measured $\boldsymbol{\mu}$-D signatures. 
	%Accordingly, the dataset build up is strongly based on realizing the outline and detailed structure of the gait cycle for each collected subject.	
	\vspace{-2mm}
	\subsection{Radar System Parametrization}
	For the proposed experiment, a CW radar operating at a carrier frequency of $f_c =$ \unit[25]{GHz} is used. The main focus of this study is the analysis of the velocity of different body parts. Thus, no range information is required and the $\boldsymbol{\mu}$-D signature is extracted without any frequency modulation. The radar parametrization considered for our experiment is shown in Table~\ref{tab:param}. The velocity attributes are calculated based on equations presented in \cite{microdopplerBook}. 
	% based on equations presented in \cite{microdopplerBook}.
	\begin{table}
		\centering
		\caption{The used parameters for $\boldsymbol{\mu}$-D  acquisition.\label{tab:param}}
		\resizebox{\columnwidth}{!}{
			\begin{tabular}{l r l r}
				\toprule
				\multicolumn{2}{c}{Radar Parametrization} & \multicolumn{2}{c}{Velocity Attributes} \\
				\midrule
				Center frequency  & \unit[25]{GHz} & {} & {}\\
				Sampling frequency   & \unit[128]{kHz} & Maximum velocity & \unit[6]{m/s}\\
				Pulse frequency   & \unit[2]{kHz} & Velocity resolution& \unit[2]{cm/s}\\
				Chirp duration   & \unit[1]{ms} & {} & {}\\
				\bottomrule
			\end{tabular}
		}
%		\vspace{-2mm}
	\end{table}
		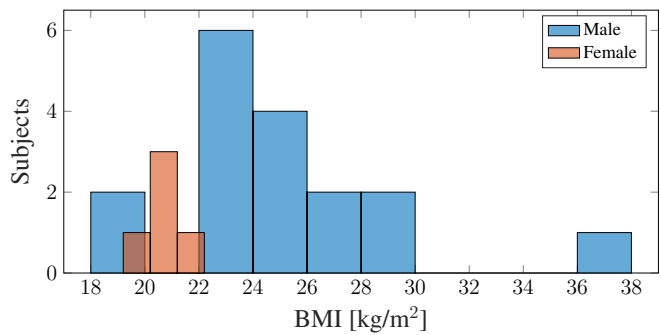
\begin{figure}
	\centering
	\resizebox{1\columnwidth}{!}{
		\input{bmiDist.tex}
	}
	\caption{The distribution of both genders participating in the experiment based on the BMI. Variations from 18.5 to 37 kg/m$^2$ with an overall mean for both genders of 24 kg/m$^2$. \label{fig:bmiDist}}
	\vspace{-8mm}
\end{figure}
   
    \par When studying the $\boldsymbol{\mu}$-D signature of a non-rigid body motion such as a walking human, the signature of each subject is expected to be unique due to the different walking style and detected radar cross section (RCS) \cite{Dorp2003RCS}. By definition, a $\boldsymbol{\mu}$-D signature of a human gait is analyzed as the superposition of the signatures due to the individual motions of different body parts such as arms, legs and feet. Accordingly, the backscattered radar signal would have many superimposed frequency components with significant temporal variations. To visualize the variations in frequency and thus velocity, a time-frequency analysis using the short-time Fourier transform (STFT) is required. Based on Eq.~\ref{eq:stft}, STFT can be described as the Fourier transform $Y(\omega,\tau)$ of a specified overlapping sliding window function $w(t-\tau)$ with a certain window size moving over the time signal $y(t)$. Typically, a window size of 512 samples with an overlap of 75\% between consecutive windows is proposed. To obtain a velocity resolution of \unit[2]{cm/s} and a temporal resolution of \unit[4]{ms}. Finally, the logarithmic squared magnitude of $Y(\omega,\tau)$ is used to visualize the $\boldsymbol{\mu}$-D signature as a \textit{time-frequency-power} distribution. 
	\begin{equation}
	Y(\omega,\tau) = \int_{-\infty}^{\infty}w(t-\tau)y(t)e^{-j\omega t} dt \label{eq:stft}
	\end{equation}
	\subsection{Experimental Setup}
	\begin{figure}
		\centering
		\includegraphics[scale=0.35]{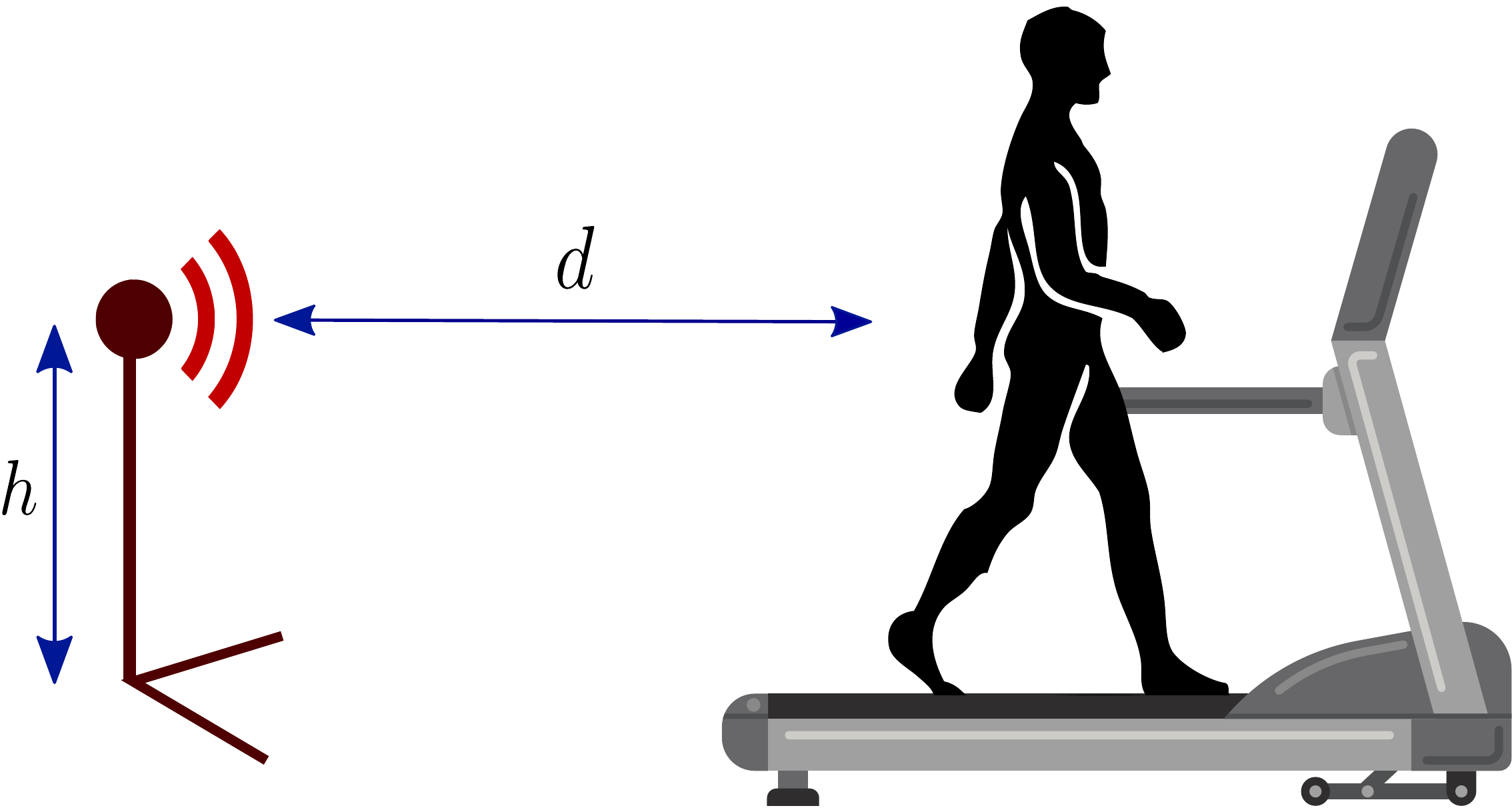}
		\caption{The experiment setup for dataset collection. 22 subjects walk on a treadmill, while keeping the radar at a constant height $h$ and distance $d$.}
		\label{fig:treadmill}
		\vspace{-12mm}
	\end{figure}
			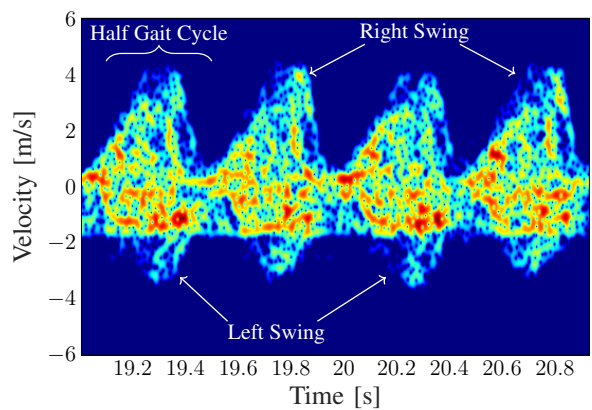
\begin{figure}
		\centering
		\resizebox{1\columnwidth}{!}{
			\input{gaitMD.tex}
		}
		\caption{The $\boldsymbol{\mu}$-D signature of two full gait cycles. It can be seen that the left swing half gaits are very similar and the same holds for the right swings. \label{fig:gMD}}
		\vspace{-4mm}
	\end{figure}
	For the required study, a dataset covering varying human body characteristics is required. Thus, 22 subjects of different genders, weights and heights are considered. The subjects are divided as 5 females and 17 males with weights from \unit[54]{kg} to \unit[115]{kg} and heights from \unit[1.62]{m} to \unit[1.95]{m}. To jointly monitor the variation in height and weight, a body mass index (BMI) is calculated for each subject as weight/height$^2$. However, BMI can sometimes be misleading in measuring the volume of the subject as it does not take into account the bone density, body fat, and muscle mass \cite{nuttall2015body}. To mitigate such effect, the participants are selected such that their BMIs are directly correlated with their body volumes. The BMI distribution of selected subjects is shown in Fig.~\ref{fig:bmiDist}.
	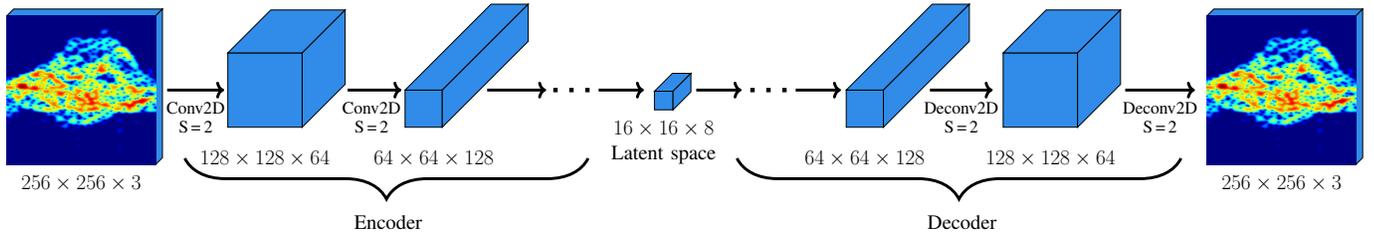
\begin{figure*}
		\centering
		\resizebox{18.5cm}{!}{\input{CAE.tikz}}
		\caption{The proposed convolutional autoencoder. The 256$\times$256$\times$3 input image is encoded to a latent space of 16$\times$16$\times$8 and then decoded back to the same high dimensional input space. \label{fig:AE}}
		\vspace{-12mm}
	\end{figure*}
		\begin{figure*}
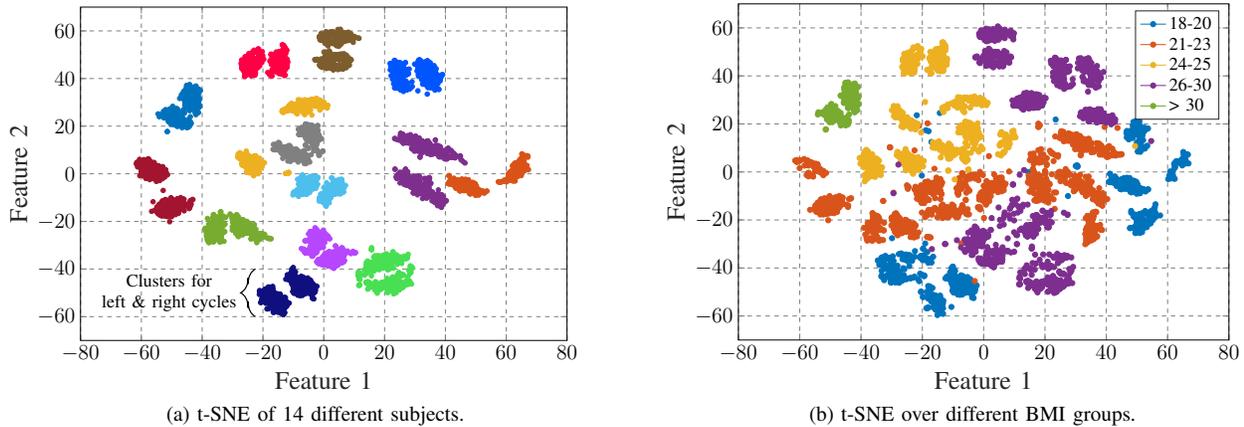

	\centering
	\centerline{\subfloat[t-SNE of 14 different subjects.]{\resizebox{.95\columnwidth}{!}{\input{tsneCycles2.tex}}\label{fig:t1}}
		\hspace{2mm}
		\subfloat[t-SNE over different BMI groups.]{\resizebox{.95\columnwidth}{!}{\input{tsne2.tex}}\label{fig:t2}}}
	\caption{t-SNE of the encoded latent space for the collected half gait $\boldsymbol{\mu}$-D signatures. (a) t-SNE for 14 subjects, where two clusters for each subject are observed (left and right swing). (b) t-SNE for all 22 subjects divided into 5 BMI groups.\label{fig:tsne}}
	\vspace{-3mm}
\end{figure*}

	The radar is placed at a certain height and a distance from the back of a treadmill as shown in Fig.~\ref{fig:treadmill}. The subjects are asked to walk on the treadmill away from the radar with an average velocity of \unit[1.6]{m/s}. Both the height $h$ and the distance $d$ are fixed for all subjects as \unit[1]{m} and \unit[3]{m}, respectively. By doing this, the only variable in the experiment is the walking subject. Therefore, the $\boldsymbol{\mu}$-D signature should only depend on the walking style and the RCS of the measured subject. It is expected that the RCS of a subject is related to its cross section and thus BMI.
	
	 A $\boldsymbol{\mu}$-D signature of a full gait cycle is interpreted as two main half cycles. Each half gait represents the changing velocity of different body parts. For instance, the left feet induces the highest velocity component within one-half gait compared to the left leg and the right arm. These effects are reversed in the following half gait, while the periodicity of the full gait cycle is preserved as shown in Fig.~\ref{fig:gMD}. Due to the periodicity of each half gait cycle, the $\boldsymbol{\mu}$-D signature is sliced on half gait basis to ensure that the data is reflecting the characteristics of the walking motion. The experiment acquisition time for each subject is \unit[180]{s}. Depending on the height of a subject, the number of steps per second can vary. However, an average of \unit[0.5]{s} for a half gait can be considered. Accordingly, an average amount of 360 half gait cycles per subject is recorded. Finally, each half gait cycle is saved as an unsigned 16 bits integer RGB image with a size of 256$\times$256$\times$3 to be used as an input for the proposed networks.

	\section{Autoencoder for Signature Analysis \label{sec:unsupervised}}  %% CAE - TSNE
	From the designed experimental setup, a total of 7920 half gait $\boldsymbol{\mu}$-D images were collected. In each image, a significant amount of useful information regarding the motion characteristics can be extracted. To study the effect of the human body characteristics on the $\boldsymbol{\mu}$-D signature, an unsupervised autoencoder is employed to reduce the dimensionality of the input data to a certain latent space. Since the input $\boldsymbol{\mu}$-D signature is represented in the \textit{time-frequency} space, a convolutional autoencoder (CAE) is used to extract the latent features from the images.
	
	A CAE has an encoder-decoder structure \cite{le2013CAE}. \textbf{Encoder} downsamples a high dimensional input $x$ into a latent space representation $z$ or bottleneck (maximum compression point) by a sequence of convolutional layers. \textbf{Decoder} upsamples back from the latent space representation $z$ to reconstruct the high dimensional input $\hat{x}$ using a sequence of deconvolutional layers. The main aspect of the CAE is to minimize the cost function represented as the mean square error between the input $x$ and the output $\hat{x}$. As shown in Fig.~\ref{fig:AE}, the proposed CAE takes an input image of 256$\times$256$\times$3 and then a 4-layer convolutional network with a stride of 2 is used to reduce the input to a latent space of 16$\times$16$\times$8$\,=\,$2048. A mirrored deconvolutional network is applied to reconstruct the input at the decoder side. Therefore, the extracted 2048 feature vector from the downsampled latent space represents the vital information in the half gait $\boldsymbol{\mu}$-D signature. This exact feature space size is obtained through hyper-parameters optimization of our proposed CAE. Accordingly, further dimension reduction using extra layers can lead to an imperfect reconstruction of the input image and thus loss of vital information in the bottle neck latent space.
	
	To visualize the distribution of the learned latent space representation, a further dimension reduction is applied using the t-distributed stochastic neighbor embedding (t-SNE) introduced in \cite{maaten2008visualizing}. t-SNE is a nonlinear dimension reduction technique that can be used to visualize the distribution of the encoded latent space of size 2048 in a two-dimensional (2D) space. The main idea behind t-SNE is to construct two joint probability distributions for both the high dimensional and the mapped low dimensional space. Finally, the Kullback–Leibler divergence between both distributions is minimized. Therefore, high dimensional samples with similar global or local structures are highly probable to be visualized as neighboring points in the 2D space.
	
	\begin{figure*}
		\centering
		\includegraphics[width=1\textwidth]{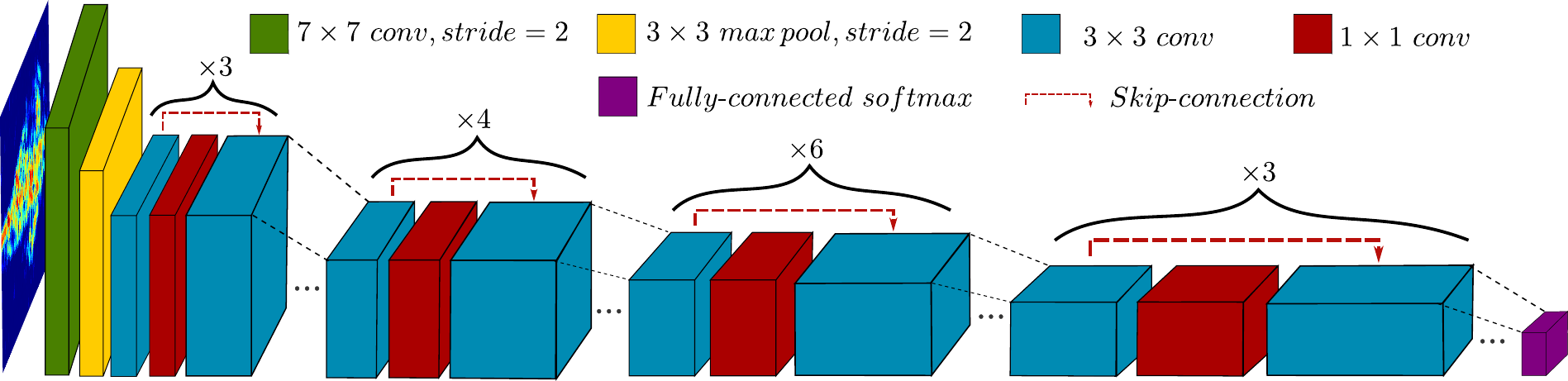}
		\caption{Architecture of ResNet-50 used for human identification. Downsampling by a stride of 2 is applied before each residual block. Re-LU activation is used for all layers except softmax for the output layer. \label{fig:netArch}}
		\vspace{-3.5mm}
	\end{figure*}
	
	Based on this approach, we can reduce the dimensions of the encoded half gait signatures into a 2D space that can be easily interpreted. In Fig.~\ref{fig:t1}, the reduced t-SNE is plotted for 14 subjects out of 22 for the sake of easier visualization. It can be observed that t-SNE can identify each subject as two close, but separate clusters for the left and right swing cycles. This verifies the strength of t-SNE as it can identify each half gait as separate clusters within the same area occupied by the subject in the 2D space. Moreover, neighboring clusters of two different subjects can belong to two different gender groups with different heights and weights. However, the BMI of neighboring clusters is similar in most of the cases. As shown in Fig.~\ref{fig:t2}, the 22 subjects can be divided into 5 BMI groups. Normally, clusters belonging to the same BMI group are distributed over the same area in the 2D space. Since the walking style can have a significant influence on the $\boldsymbol{\mu}$-D signature outline and local structure, some BMI groups tend to cluster over two major areas in the reduced space. It can also be observed that the latent space distribution of the only participant belonging to the obese category (BMI over 30) is visualized as a separate cluster in the 2D space.
	
	\section{Person-ID Network Architecture \label{sec:id}}
	Based on the presented study, we can deduce that subjects with similar BMI and walking styles are highly probable to have similar $\boldsymbol{\mu}$-D signatures. Thus, classifying them as different subjects needs a deeper architecture that can utilize local details in the spectrum. This was observed as authors in \cite{Cao2018ID} used a shallow 6-layer architecture to classify 20 subjects of a BMI range 18-32 kg/m$^2$, and they reached an acceptable performance of 97.1 \% with only 4 subjects. Adding more subjects causes more similar BMIs and walking styles and thus a degradation in the classification performance to 68.9\% for 20 participants. 
	
	Accordingly, a deeper architecture is used in this paper to classify the 22 participants based on the collected half gait $\boldsymbol{\mu}$-D signatures. Since the real-time aspect is an important issue, an architecture with fewer computations and short inference time is required. Thus, the ResNet-50 architecture \cite{He_2016_CVPR} is used as its number of parameters and operations is much lower than that of VGG-19 \cite{SimonyanZ14a}. As shown in Fig.~\ref{fig:netArch}, ResNet-50 architecture is based on residual blocks, where a skip connection is applied every second layer:
	\begin{equation}
	a[l+2]=relu(a[l]+z[l+2]) \label{eq:res}
	\end{equation}
	where $l$ is the layer index, $a[.]$ is the layer activation and $z[.]$ is the output of the layer before $relu$ activation. These connections allow the training of very deep networks without degradation in performance. Which can be explained as in case of low weights in the output $z[l+2]$, the network learns the identity mapping of the activation $a[l]$. In other cases, the network can learn weights and biases that can improve the performance. Thus, the skip connections will either stabilize or improve the performance of the network. 
	
	\par The skip connections are valid in case of dimensions agreement which is not always the case. To resolve this issue, a zero padding is used to allow the residual operation over these layers. Moreover, it is known that very deep networks can suffer from vanishing gradient problems and this can be mitigated by using batch normalization \cite{batch2015}. Finally, the output of the last convolutional layer is flattened and passed to a fully connected layer. Then a softmax layer generates the probability of 22 classes corresponding to the participants.
	 
	 \section{Results}
	 	\begin{figure*}
	 	\centering
	 	\centerline{
	 		\subfloat[High SNR]{
	 			\begin{overpic}[width=0.18\textwidth]%
	 				{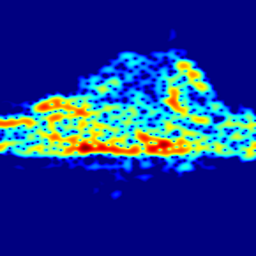}
	 				\centering
	 				\put(24,98){BMI: 23.66}
	 			\end{overpic}
	 			\hspace{2mm}
	 			\begin{overpic}[width=0.18\textwidth]%
	 				{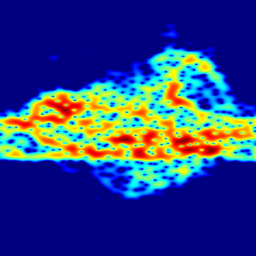}
	 				\centering
	 				\put(24,98){BMI: 37.55}
	 			\end{overpic}
	 			\label{fig:cl1}}
	 		\hspace{20mm}
	 		\subfloat[Low SNR]{
	 			\begin{overpic}[width=0.18\textwidth]%
	 				{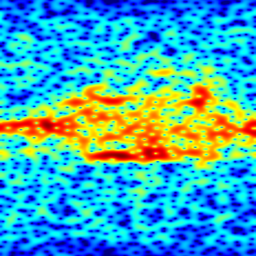}
	 				\centering
	 				\put(24,98){BMI: 18.59}
	 			\end{overpic}
	 			\hspace{2mm}
	 			\begin{overpic}[width=0.18\textwidth]%
	 				{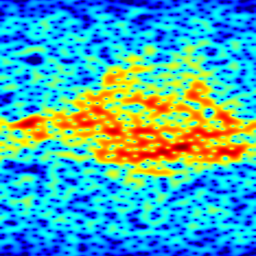}
	 				\centering
	 				\put(24,98){BMI: 21.93}
	 			\end{overpic}
	 			\label{fig:n1}}
	 	}
	 	\caption{Examples of different half gait $\boldsymbol{\mu}$-D signatures collected from subjects with different BMIs. (a) The radar is placed at a distance $d=\,$\unit[3]{m}, i.e., high SNR experiment. (b) The radar is placed at a distance $d=\,$\unit[10]{m}, i.e., low SNR experiment.\label{fig:exps}}
	 	\vspace{-7mm}
	 \end{figure*}
 	 \begin{figure*}
 	\centering
 	\centerline{\hspace{8mm}
 		\captionsetup[subfigure]{oneside,margin={-0.9cm,0cm}}
 		\subfloat[Confusion on high SNR dataset]{
			 	\resizebox{1\columnwidth}{!}{
				\input{confmatClean.tex}
			}
 			\label{fig:cmClean}}
 		\hspace{0.5mm}
 		\subfloat[Confusion on varying SNR dataset]{
		\resizebox{1\columnwidth}{!}{
	\input{confmatMixed.tex}
}
 			\label{fig:cmNoisy}}
 	}
 	\caption{(a) Confusion matrix for $\boldsymbol{\mu}$-D signatures collected at high SNR. An overall accuracy of 98\% is achieved. (b) Confusion matrix for $\boldsymbol{\mu}$-D signatures collected at varying SNR. An overall accuracy of 84\% is achieved. \label{fig:cm}}
 	\vspace{-3mm}
 \end{figure*}
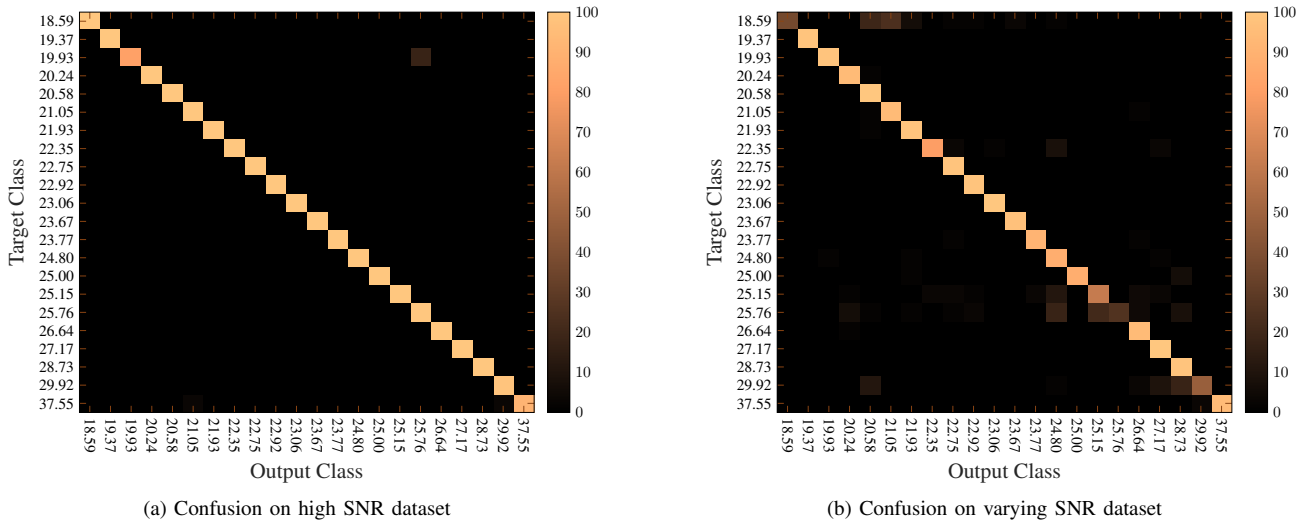

	  The ResNet-50 architecture is trained over the 7920 labeled half gait $\boldsymbol{\mu}$-D signatures using Adam optimizer. The model took about half an hour to train on a single NIVIDIA Titan X GPU. Examples of collected signatures from different BMI groups are shown in Fig.~\ref{fig:cl1}. Subjects with low BMI tend to have a lower power inside the signature local structure due a relatively lower RCS. The global structure of the signature on the other hand, is mainly related to the swinging limbs velocity components and so to the walking style.
	  
	  To evaluate the ResNet-50 performance, another experiment is conducted on the same 22 subjects with the same treadmill velocity and experiment duration to collect an additional set of 7920 signatures for testing. The radar is placed at the same distance of \unit[3]{m} from the treadmill. The network achieved an overall accuracy of 98\% on the unseen test set. The model inference time, i.e., duration to classify one half gait signature is only \unit[200]{ms}.
	  
	   A normalized confusion matrix for the collected test set is shown in Fig.~\ref{fig:cmClean}. It can be concluded that adding more layers with the proposed ResNet-50 architecture increased the classification performance for multiple subjects (>20), though most of the participants belong to the same BMI group  21-26 kg/m$^2$ with similar walking styles.
	   
	  To validate the hypothesis shown in Sec.~\ref{sec:unsupervised} that the BMI of a subject can affect his signature, a more realistic person identification experiment is applied. In this experiment, the radar is moved to a larger distance of \unit[10]{m} from the treadmill. Increasing the target range from the radar leads to a reduction in the SNR based on equations presented in \cite{richards2005fundamentals}. This is mainly observed for subjects with lower BMIs due to the lower detected RCS. Moreover, the  $\boldsymbol{\mu}$-D signatures of subjects with comparable BMI have lost partially their unique global and local structures. Thus, the classification task is expected to be more challenging due to a higher similarity between the half gait signatures as shown in Fig.~\ref{fig:n1}.   
	  
	Based on this setup, an additional dataset is collected for all 22 subjects to analyze the lower SNR effect. All other experiment conditions including the walking velocity and data acquisition time are the same. The low-SNR dataset is combined with the high-SNR dataset and then divided into train and test sets. Accordingly, the combined dataset contains more images with different SNR values and hence it is more challenging for person identification. The SNR variation forces the network to learn more sophisticated local $\boldsymbol{\mu}$-D structures to distinguish between different subjects. The same network structure from Sec.~\ref{sec:id} was trained by the mixed-SNR training set. It achieves an accuracy of 84\% on the unseen test set. As shown in Fig.\ref{fig:cmNoisy}, the highest confusion is mostly between subjects with similar BMI values.

	 \section{Conclusion}
	 In this paper, an experimental study is presented to illustrate the effect of human body characteristics on the $\boldsymbol{\mu}$-D signatures of walking persons. A CW radar is used to measure the $\boldsymbol{\mu}$-D signatures of 22 subjects with different genders and body structures walking on a treadmill. Accordingly, the main attributes affecting the measured $\boldsymbol{\mu}$-D signatures are the RCS and walking style of the moving subject. The BMI is used as a metric for the body volume of the selected participants ranging from slim bodies at 18-20 kg/m$^2$ till obese levels at 37 kg/m$^2$. A CAE is used to encode the half gait $\boldsymbol{\mu}$-D signatures into a latent space representation. A nonlinear dimension reduction based on t-SNE is applied to visualize the latent space distribution in 2D. Based on visual interpretation, the BMI of the subjects as well as the walking styles are found to have a direct influence on the corresponding $\boldsymbol{\mu}$-D signatures. A deep ResNet-50 architecture is trained for person identification based on the $\boldsymbol{\mu}$-D signatures. The architecture correctly identifies the $\boldsymbol{\mu}$-D signatures with an accuracy of 98\% on low SNR values and 84\% on different SNR values with confusions mostly in subjects with similar BMIs.
	 
	 The proposed experiments included only one motion aspect angle (radial). For future work, signatures from different aspect angles should be taken into consideration. For more realistic use case, examples of normal walking in free space without treadmill should be included in the study and the person identification network. Moreover, a study of how accurate an image based regression network can be in estimating the BMI of a subject from the collected $\boldsymbol{\mu}$-D signature will be interesting to consider.

	\bibliographystyle{IEEEtran}
	
	%\begin{thebibliography}{1}
	%\end{thebibliography}
	
%	\bibliography{refs1}

\end{document}

%% file: bmiDist.tex
\definecolor{mycolor1}{rgb}{0.00000,0.44700,0.74100}%
\definecolor{mycolor2}{rgb}{0.85000,0.32500,0.09800}%
\begin{tikzpicture}

\begin{axis}[%
width=4.521in,
height=2in,
at={(0.758in,0.488in)},
scale only axis,
xmin=17,
xmax=39,
xlabel style={font=\color{white!15!black}},
xlabel={\Large BMI [kg/m$^2$]},
ymin=0,
ymax=6.5,
yticklabel style = {font=\large},
xticklabel style = {font=\large},
ylabel style={font=\color{white!15!black}},
ylabel={\Large Subjects},
axis background/.style={fill=white},
legend style={legend cell align=left, align=left, draw=white!15!black}
]
\addplot[ybar interval, fill=mycolor1, fill opacity=0.6, draw=black, area legend] table[row sep=crcr] {%
	x	y\\
	18	2\\
	20	0\\
	22	6\\
	24	4\\
	26	2\\
	28	2\\
	30	0\\
	32	0\\
	34	0\\
	36	1\\
	38	1\\
};
\addlegendentry{Male}

\addplot[ybar interval, fill=mycolor2, fill opacity=0.6, draw=black, area legend] table[row sep=crcr] {%
	x	y\\
	19.2	1\\
	20.2	3\\
	21.2	1\\
	22.2	1\\
};
\addlegendentry{Female}

\end{axis}
\end{tikzpicture}%

%% file: gaitMD.tex
\begin{tikzpicture}

\begin{axis}[%
width=4.521in,
height=3in,
at={(0.758in,0.481in)},
scale only axis,
point meta min=-23.1839750284342,
point meta max=16.8160249715658,
axis on top,
xmin=19.0064224400436,
xmax=20.9304652041519,
xlabel style={font=\color{white!15!black}},
xlabel={\LARGE Time [s]},
ymin=-6.00758272,
ymax=6.00758272,
ytick style={draw=none},
xtick style={draw=none},
yticklabel style = {font=\Large},
xticklabel style = {font=\Large},
ylabel style={font=\color{white!15!black}},
ylabel={\LARGE Velocity [m/s]},
axis background/.style={fill=white},
legend style={legend cell align=left, align=left, draw=white!15!black}
]
\addplot [forget plot] graphics [xmin=19.0064224400436, xmax=20.9304652041519, ymin=-6.00758272, ymax=6.00758272] {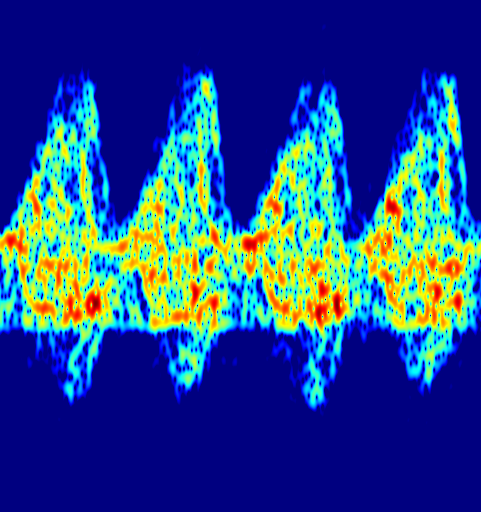};
\draw [white,thick,decoration={brace,amplitude=10pt,raise=5pt},decorate] 
(axis cs:19.1,4.2) --
node[above=12pt] {\color{white} \Large Half Gait Cycle} 
(axis cs:19.5,4.2);
\draw [white, line width=0.3mm, ->](axis cs:20.08,5.081) -- (axis cs:19.87,4.025);
\draw [white, line width=0.3mm, ->](axis cs:20.44,5.081) -- (axis cs:20.65,4.025);
\node[right,align=left, font=\color{white}]
at (axis cs:20.04,5.47) {\Large Right Swing};
\draw [white, line width=0.3mm, ->](axis cs:19.56,-4.799) -- (axis cs:19.38,-3.227);
\draw [white, line width=0.3mm, ->](axis cs:19.9,-4.799) -- (axis cs:20.16,-3.227);
\node[left,align=left, font=\color{white}]
at (axis cs:19.95,-5.24) {\Large Left Swing};
%(axis cs:19.1,4.2) --
%node[above=12pt] {\color{white} Half Gait Cycle} 
%(axis cs:19.5,4.2);
\end{axis}

\begin{axis}[%
width=5.833in,
height=4.375in,
at={(0in,0in)},
scale only axis,
xmin=0,
xmax=1,
ymin=0,
ymax=1,
axis line style={draw=none},
ticks=none,
axis x line*=bottom,
axis y line*=left,
legend style={legend cell align=left, align=left, draw=white!15!black}
]

%\draw [decorate,line width=1mm,decoration={brace,amplitude=40pt,mirror,raise=4pt},yshift=10pt]
%(19.1,5) -- (19.5,5) node [black,midway,below,xshift=0.8cm] {};
%\node at (19.25,5) {\huge Encoder};
%\node[below right, align=left, font=\bfseries\color{white}]
%at (rel axis cs:0.149,0.85) {Half Gait Cycle};
\end{axis}
\end{tikzpicture}%

%% file: CAE.tikz
\newcommand{\w}{20}
\newcommand{\dist}{7}

\newcommand{\distEinsZwei}{8}
\newcommand{\distEinsDrei}{\distEinsZwei+6}
\newcommand{\distEinsDreiB}{\distEinsDrei+6}

\newcommand{\distEinsVier}{\distEinsDreiB+5}

\definecolor{cl}{rgb}{0.19, 0.55, 0.91}
\definecolor{red}{rgb}{0,0,0}
%\newcommand{\cl}{blue}
%\begin{document}
\begin{tikzpicture}
\tikzstyle{dotted}= [dash pattern=on \pgflinewidth off 4mm] 

%\node[parallelepiped,draw=red,fill=yellow,
%minimum width=2.5cm,minimum height=1.5cm] (1) {Node One};
%\node[parallelepiped,draw=blue,fill=green,
%minimum height=2.5cm,minimum width=1.5cm,parallelepiped offset x=7mm] (2)
%at (6,0) {Node Two};
%\draw[ultra thick, ->] (1) -- (2);
%\node (0,0) (12) {Node};

\newcommand{\ah}{0.256*\w/2}

% Pfeil 1
\node[name=PfeilEinsAnf] at (0.245*\w*1.1,\ah) {};
\node[name=PfeilEinsEnd] at (\distEinsZwei*0.95,\ah) {};
\draw[line width=1mm, ->] (PfeilEinsAnf) -- (PfeilEinsEnd) node [below=0.7em,midway] {\fontsize{17}{17}\selectfont Conv2D} node [below=2.5em,midway] {\fontsize{17}{17}\selectfont S\,=\,2};

% Pfeil 2 ends at \distEinsDrei at begin at distEinsZwei + 0.128*\w
\node[name=PfeilZweiAnf] at (\distEinsZwei*1.45, \ah) {};
\node[name=PfeilZweiEnd] at (\distEinsDrei*0.95, \ah) {};
\draw[line width=1mm, ->] (PfeilZweiAnf) -- (PfeilZweiEnd) node [below=0.7em,midway] {\fontsize{17}{17}\selectfont Conv2D \hspace{1mm}} node [below=2.5em,midway] {\fontsize{17}{17}\selectfont S\,=\,2 \hspace{1mm}};

% Pfeil 3 
\node[name=Pfeil3Anf] at (\distEinsDrei*1.4, \ah) {};
\node[name=Pfeil3End] at (\distEinsDreiB*0.78, \ah) {};
\draw[line width=1mm, ->] (Pfeil3Anf) -- (Pfeil3End);

% Dots Encoding
%\node[name=dots] at (\distEinsDreiB*1.15, \ah) {\ldots};
\draw[dotted,line width = 1.5mm] (\distEinsDreiB*0.8, \ah) -- (\distEinsDreiB*1.02, \ah);

\draw[line width=1mm, ->] (\distEinsDreiB*1.06, \ah) -- (\distEinsDreiB*1.32, \ah);

\draw[line width=1mm, ->] (\distEinsDreiB*1.62, \ah) -- (\distEinsDreiB*1.88, \ah);

\draw[dotted,line width = 1.5mm] (\distEinsDreiB*1.93, \ah) -- (\distEinsDreiB*2.15, \ah);

\draw[line width=1mm, ->] (\distEinsDreiB*2.19, \ah) -- (\distEinsDreiB*2.45, \ah);

\draw[line width=1mm, ->] (\distEinsDreiB*2.96, \ah) -- (\distEinsDreiB*3.32, \ah) node [below=0.7em,midway] {\fontsize{17}{17}\selectfont Deconv2D} node [below=2.5em,midway] {\fontsize{17}{17}\selectfont S\,=\,2};;

\draw[line width=1mm, ->] (\distEinsDreiB*4.08, \ah) -- (\distEinsDreiB*4.48, \ah) node [below=0.7em,midway] {\fontsize{17}{17}\selectfont Deconv2D} node [below=2.5em,midway] {\fontsize{17}{17}\selectfont S\,=\,2};;
%-- node[auto=false]{} (2);
% Pfeil 4 
%\node[name=Pfeil4Anf] at (\distEinsDrei*1.4, \ah) {};
%\node[name=Pfeil4End] at (\distEinsVier, \ah) {};
%\draw[ultra thick, ->] (Pfeil4Anf) -- (Pfeil4End);

% 256x256x3 INPUT
	
	[every edge quotes/.append style={auto, text=blue}]
	\pgfmathsetmacro{\xInp}{0.256*\w}
	\pgfmathsetmacro{\y}{0.256*\w}
	\pgfmathsetmacro{\z}{0.03*\w}
	
	\draw[red,fill=cl] (0,0,0) coordinate (a) --++ (\xInp,0,0) coordinate (b) --++ (0,\y,0) coordinate (c) --++ (-\xInp,0,0) coordinate (d) --cycle;
	\draw[red, fill=cl] (d) -- (c) --++ (0,0,-\z) coordinate (g) --++ (-\xInp, 0, 0) coordinate (h) -- cycle;
	\draw[red, fill=cl] (b) -- (c) -- (g) --++ (0,-\y, 0) coordinate (f) -- cycle;

	\node[name,anchor=center,opacity=1] at (\xInp/2, \y/2) {\includegraphics[width=\xInp cm,height=\y cm]{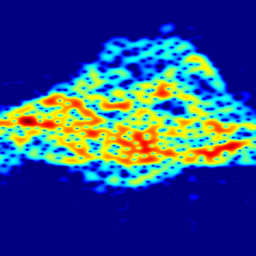}};
	
	\node at (\xInp/2, -\y/8) {\huge $256\times256\times3$};

% 128x128x64  EINS

[every edge quotes/.append style={auto, text=blue}]
\pgfmathsetmacro{\x}{0.128*\w}
\pgfmathsetmacro{\y}{0.128*\w}
\pgfmathsetmacro{\z}{0.064*3*\w}

\draw[red,fill=cl] (\distEinsZwei*0.95,\y/2,0) coordinate (EINSa) --++ (\x,0,0) coordinate (b) --++ (0,\y,0) coordinate (c) --++ (-\x,0,0) coordinate (d) --cycle;
\draw[red, fill=cl] (d) -- (c) --++ (0,0,-\z) coordinate (g) --++ (-\x, 0, 0) coordinate (h) -- cycle;
\draw[red, fill=cl] (b) -- (c) -- (g) --++ (0,-\y, 0) coordinate (f) -- cycle;	

\node[below=0.1em] at ($ (EINSa) + (\x/2, -\y/4) $) {\huge $128\times128\times64$};

% 64x64x128 ZWEI 

[every edge quotes/.append style={auto, text=blue}]
\pgfmathsetmacro{\xZWEI}{0.064*\w}
\pgfmathsetmacro{\yZWEI}{0.064*\w}
\pgfmathsetmacro{\z}{0.128*3*\w}

\draw[red,fill=cl] (\distEinsDrei*0.95,\yZWEI,0) coordinate (ZWEIa) --++ (\xZWEI,0,0) coordinate (b) --++ (0,\yZWEI,0) coordinate (c) --++ (-\xZWEI,0,0) coordinate (d) --cycle;
\draw[red, fill=cl] (d) -- (c) --++ (0,0,-\z) coordinate (g) --++ (-\xZWEI, 0, 0) coordinate (h) -- cycle;
\draw[red, fill=cl] (b) -- (c) -- (g) --++ (0,-\yZWEI, 0) coordinate (f) -- cycle;

\node[below=0.1em] at ($ (ZWEIa) + (\xZWEI/2, -\yZWEI/2) $) {\hspace{7mm}\huge $64\times64\times128$};

% 16x16x8 TINY

\newcommand{\temp}{2}

[every edge quotes/.append style={auto, text=blue}]
\pgfmathsetmacro{\xLAT}{0.016*\w*\temp}
\pgfmathsetmacro{\yLAT}{0.016*\w*\temp}
\pgfmathsetmacro{\z}{0.008*10*\w}

\draw[red,fill=cl] (\distEinsDreiB*1.38,0.094*\w,0) coordinate (LATa) --++ (\xLAT,0,0) coordinate (b) --++ (0,\yLAT,0) coordinate (c) --++ (-\xLAT,0,0) coordinate (d) --cycle;
\draw[red, fill=cl] (d) -- (c) --++ (0,0,-\z) coordinate (g) --++ (-\xLAT, 0, 0) coordinate (h) -- cycle;
\draw[red, fill=cl] (b) -- (c) -- (g) --++ (0,-\yLAT, 0) coordinate (f) -- cycle;

\node at ($ (LATa) + (\xLAT/2, -\yLAT) $) {\huge $16\times16\times8$};

\node at ($ (LATa) + (\xLAT/2, -\yLAT-0.9) $) {\huge Latent space};
%64x64x128

[every edge quotes/.append style={auto, text=blue}]
\pgfmathsetmacro{\xDEC}{0.064*\w}
\pgfmathsetmacro{\yDEC}{0.064*\w}
\pgfmathsetmacro{\z}{0.128*3*\w}

\draw[red,fill=cl] (\distEinsDreiB*2.48,\yDEC,0) coordinate (DECa) --++ (\xDEC,0,0) coordinate (b) --++ (0,\yDEC,0) coordinate (c) --++ (-\xDEC,0,0) coordinate (d) --cycle;
\draw[red, fill=cl] (d) -- (c) --++ (0,0,-\z) coordinate (g) --++ (-\xDEC, 0, 0) coordinate (h) -- cycle;
\draw[red, fill=cl] (b) -- (c) -- (g) --++ (0,-\yDEC, 0) coordinate (f) -- cycle;

\node[below=0.1em] at ($ (DECa) + (\xDEC/2, -\yDEC/2) $) {\huge $64\times64\times128$};

% 128x128x64 vorletzter

[every edge quotes/.append style={auto, text=blue}]
\pgfmathsetmacro{\xVL}{0.128*\w}
\pgfmathsetmacro{\yVL}{0.128*\w}
\pgfmathsetmacro{\z}{0.064*3*\w}

\draw[red,fill=cl] (\distEinsDreiB*3.38,\yVL/2,0) coordinate (aVL) --++ (\xVL,0,0) coordinate (b) --++ (0,\yVL,0) coordinate (c) --++ (-\xVL,0,0) coordinate (d) --cycle;
\draw[red, fill=cl] (d) -- (c) --++ (0,0,-\z) coordinate (g) --++ (-\xVL, 0, 0) coordinate (h) -- cycle;
\draw[red, fill=cl] (b) -- (c) -- (g) --++ (0,-\yVL, 0) coordinate (f) -- cycle;	

\node[below=0.1em] at ($ (aVL) +(\xVL/2, -\yDEC/2)$) {\hspace{7mm}\huge $128\times128\times64$};

% 256x256x3

[every edge quotes/.append style={auto, text=blue}]
\pgfmathsetmacro{\xOUT}{0.256*\w}
\pgfmathsetmacro{\yOUT}{0.256*\w}
\pgfmathsetmacro{\z}{0.03*\w}

\draw[red,fill=cl] (\distEinsDreiB*4.55,0,0) coordinate (OUTa) --++ (\xOUT,0,0) coordinate (b) --++ (0,\yOUT,0) coordinate (c) --++ (-\xOUT,0,0) coordinate (d) --cycle;
\draw[red, fill=cl] (d) -- (c) --++ (0,0,-\z) coordinate (g) --++ (-\xOUT, 0, 0) coordinate (h) -- cycle;
\draw[red, fill=cl] (b) -- (c) -- (g) --++ (0,-\yOUT, 0) coordinate (f) -- cycle;	

\node[,anchor=center,opacity=1]  at ($ (OUTa) +(\xOUT/2, \yOUT/2) $) {\includegraphics[width=\xOUT cm,height=\yOUT cm]{in_high_res.png}};

\node[name,anchor=center,opacity=0.3] at (\xOUT/2, \yOUT/2) {\includegraphics[width=\xOUT cm,height=\yOUT cm]{in_high_res.png}};

\node at ($ (OUTa) +(\xOUT/2, -\yOUT/8)$) {\huge $256\times256\times3$};

% ENCODER Brace
\draw [decorate,line width=1mm,decoration={brace,amplitude=40pt,mirror,raise=4pt},yshift=10pt]
(\xInp+1,0) -- (\distEinsDrei*2,0) node [black,midway,below,xshift=0.8cm] {};
\node at (\xInp+8,-2) {\huge Encoder};

\draw [decorate,line width=1mm,decoration={brace,amplitude=40pt,mirror,raise=4pt},yshift=10pt]
(\xInp+20,0) -- (\distEinsDrei*5.4,0) node [black,midway,below,xshift=0.8cm] {};
\node at (\xInp+27.75,-2) {\huge Decoder};
\end{tikzpicture}

%but what I obtain is that distances between elements are scaled but not their sizes or text size too. It's like if the picture collapses on itself, without correctly scaling (as I would imagine, like having a zoom factor)
%
%Am I missing something? Should I use a different command or what?but what I obtain is that distances between elements are scaled but not their sizes or text size too. It's like if the picture collapses on itself, without correctly scaling (as I would imagine, like having a zoom factor)
%
%Am I missing something? Should I use a different command or what?but what I obtain is that distances between elements are scaled but not their sizes or text size too. It's like if the picture collapses on itself, without correctly scaling (as I would imagine, like having a zoom factor)
%
%Am I missing something? Should I use a different command or what?

%\end{document}
	

%% file: confmatClean.tex
\begin{tikzpicture}
\definecolor{mycolor1}{rgb}{0.5451,0.2706,0.0745}
\begin{axis}[%
width=4.047in,
height=3.566in,
at={(0.679in,0.481in)},
scale only axis,
point meta min=0,
point meta max=100,
axis on top,
xmin=0.5,
xmax=22.5,
xtick={1,2,3,4,5,6,7,8,9,10,11,12,13,14,15,16,17,18,19,20,21,22},
xticklabels={{18.59},{19.37},{19.93},{20.24},{20.58},{21.05},{21.93},{22.35},{22.75},{22.92},{23.06},{23.67},{23.77},{24.80},{25.00},{25.15},{25.76},{26.64},{27.17},{28.73},{29.92},{37.55}},
xticklabel style={rotate=270},
y dir=reverse,
xlabel style={font=\color{white!15!black}},
xlabel={\Large Output Class},
ylabel style={font=\color{white!15!black}},
ylabel={\Large Target Class},
ytick style={color={mycolor1}},
xtick style={color={mycolor1}},
yticklabel style = {font=\normalsize},
xticklabel style = {font=\normalsize},
ymin=0.5,
ymax=22.5,
ytick={1,2,3,4,5,6,7,8,9,10,11,12,13,14,15,16,17,18,19,20,21,22},
yticklabels={{18.59},{19.37},{19.93},{20.24},{20.58},{21.05},{21.93},{22.35},{22.75},{22.92},{23.06},{23.67},{23.77},{24.80},{25.00},{25.15},{25.76},{26.64},{27.17},{28.73},{29.92},{37.55}},
axis background/.style={fill=white},
legend style={legend cell align=left, align=left, draw=white!15!black},
colormap={mymap}{[1pt] rgb(0pt)=(0,0,0); rgb(50pt)=(0.992063,0.62,0.394841); rgb(51pt)=(1,0.6324,0.402738); rgb(63pt)=(1,0.7812,0.4975)},
colorbar,
colorbar style={ytick style={draw=none}}
]
\addplot [forget plot] graphics [xmin=0.5, xmax=22.5, ymin=0.5, ymax=22.5] {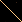};
\end{axis}

\begin{axis}[%
width=5.833in,
height=4.375in,
at={(0in,0in)},
scale only axis,
point meta min=0,
point meta max=1,
xmin=0,
xmax=1,
ymin=0,
ymax=1,
axis line style={draw=none},
ticks=none,
axis x line*=bottom,
axis y line*=left,
legend style={legend cell align=left, align=left, draw=white!15!black}
]
\end{axis}
\end{tikzpicture}%

%% file: confmatMixed.tex
\begin{tikzpicture}
\definecolor{mycolor1}{rgb}{0.5451,0.2706,0.0745}
\begin{axis}[%
width=4.047in,
height=3.566in,
at={(0.679in,0.481in)},
scale only axis,
point meta min=0,
point meta max=100,
axis on top,
xmin=0.5,
xmax=22.5,
xtick={1,2,3,4,5,6,7,8,9,10,11,12,13,14,15,16,17,18,19,20,21,22},
xticklabels={{18.59},{19.37},{19.93},{20.24},{20.58},{21.05},{21.93},{22.35},{22.75},{22.92},{23.06},{23.67},{23.77},{24.80},{25.00},{25.15},{25.76},{26.64},{27.17},{28.73},{29.92},{37.55}},
xticklabel style={rotate=270},
y dir=reverse,
xlabel style={font=\color{white!15!black}},
xlabel={\Large Output Class},
ylabel style={font=\color{white!15!black}},
ylabel={\Large Target Class},
ytick style={color={mycolor1}},
xtick style={color={mycolor1}},
yticklabel style = {font=\normalsize},
xticklabel style = {font=\normalsize},
ymin=0.5,
ymax=22.5,
ytick={1,2,3,4,5,6,7,8,9,10,11,12,13,14,15,16,17,18,19,20,21,22},
yticklabels={{18.59},{19.37},{19.93},{20.24},{20.58},{21.05},{21.93},{22.35},{22.75},{22.92},{23.06},{23.67},{23.77},{24.80},{25.00},{25.15},{25.76},{26.64},{27.17},{28.73},{29.92},{37.55}},
axis background/.style={fill=white},
legend style={legend cell align=left, align=left, draw=white!15!black},
colormap={mymap}{[1pt] rgb(0pt)=(0,0,0); rgb(50pt)=(0.992063,0.62,0.394841); rgb(51pt)=(1,0.6324,0.402738); rgb(63pt)=(1,0.7812,0.4975)},
colorbar,
colorbar style={ytick style={draw=none}}
]
\addplot [forget plot] graphics [xmin=0.5, xmax=22.5, ymin=0.5, ymax=22.5] {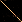};
\end{axis}

\begin{axis}[%
width=5.833in,
height=4.375in,
at={(0in,0in)},
scale only axis,
point meta min=0,
point meta max=1,
xmin=0,
xmax=1,
ymin=0,
ymax=1,
axis line style={draw=none},
ticks=none,
axis x line*=bottom,
axis y line*=left,
legend style={legend cell align=left, align=left, draw=white!15!black}
]
\end{axis}
\end{tikzpicture}%